\documentclass[10pt,twocolumn,letterpaper]{article}

\usepackage[pagenumbers]{wacv} 

\definecolor{wacvblue}{rgb}{0.21,0.49,0.74}
\usepackage[pagebackref,breaklinks,colorlinks,allcolors=wacvblue]{hyperref}

\def\wacvPaperID{794} 
\def\confName{WACV}
\def\confYear{2026}

\title{Referring Change Detection in Remote Sensing Imagery}

\author{
Yilmaz Korkmaz\textsuperscript{1} \quad
Jay N. Paranjape\textsuperscript{1} \quad
Celso M. de Melo\textsuperscript{2} \quad
Vishal M. Patel\textsuperscript{1} \\
\textsuperscript{1}Johns Hopkins University, Baltimore, USA \\
\textsuperscript{2}DEVCOM U.S. Army Research Laboratory, Adelphi, USA \\
{\tt\small \{ykorkma1, jparanj2, vpatel36\}@jhu.edu} \quad
{\tt\small celso.m.demelo.civ@army.mil}
}

\begin{document}
\maketitle
\begin{abstract}
Change detection in remote sensing imagery is essential for applications such as urban planning, environmental monitoring, and disaster management. Traditional change detection methods typically identify all changes between two temporal images without distinguishing the types of transitions, which can lead to results that may not align with specific user needs. Although semantic change detection methods have attempted to address this by categorizing changes into predefined classes, these methods rely on rigid class definitions and fixed model architectures, making it difficult to mix datasets with different label sets or reuse models across tasks, as the output channels are tightly coupled with the number and type of semantic classes. To overcome these limitations, we introduce Referring Change Detection (RCD), which leverages natural language prompts to detect specific classes of changes in remote sensing images. By integrating language understanding with visual analysis, our approach allows users to specify the exact type of change they are interested in. However, training models for RCD is challenging due to the limited availability of annotated data and severe class imbalance in existing datasets. To address this, we propose a two-stage framework consisting of (I) \textbf{RCDNet}, a cross-modal fusion network designed for referring change detection, and (II) \textbf{RCDGen}, a diffusion-based synthetic data generation pipeline that produces realistic post-change images and change maps for a specified category using only pre-change image, without relying on semantic segmentation masks and thereby significantly lowering the barrier to scalable data creation. Experiments across multiple datasets show that our framework enables scalable and targeted change detection. Project website is \href{https://yilmazkorkmaz1.github.io/RCD/}{here}.

\end{abstract}    
\section{Introduction}

\begin{figure}
    \centering
    \includegraphics[width=0.77\linewidth]{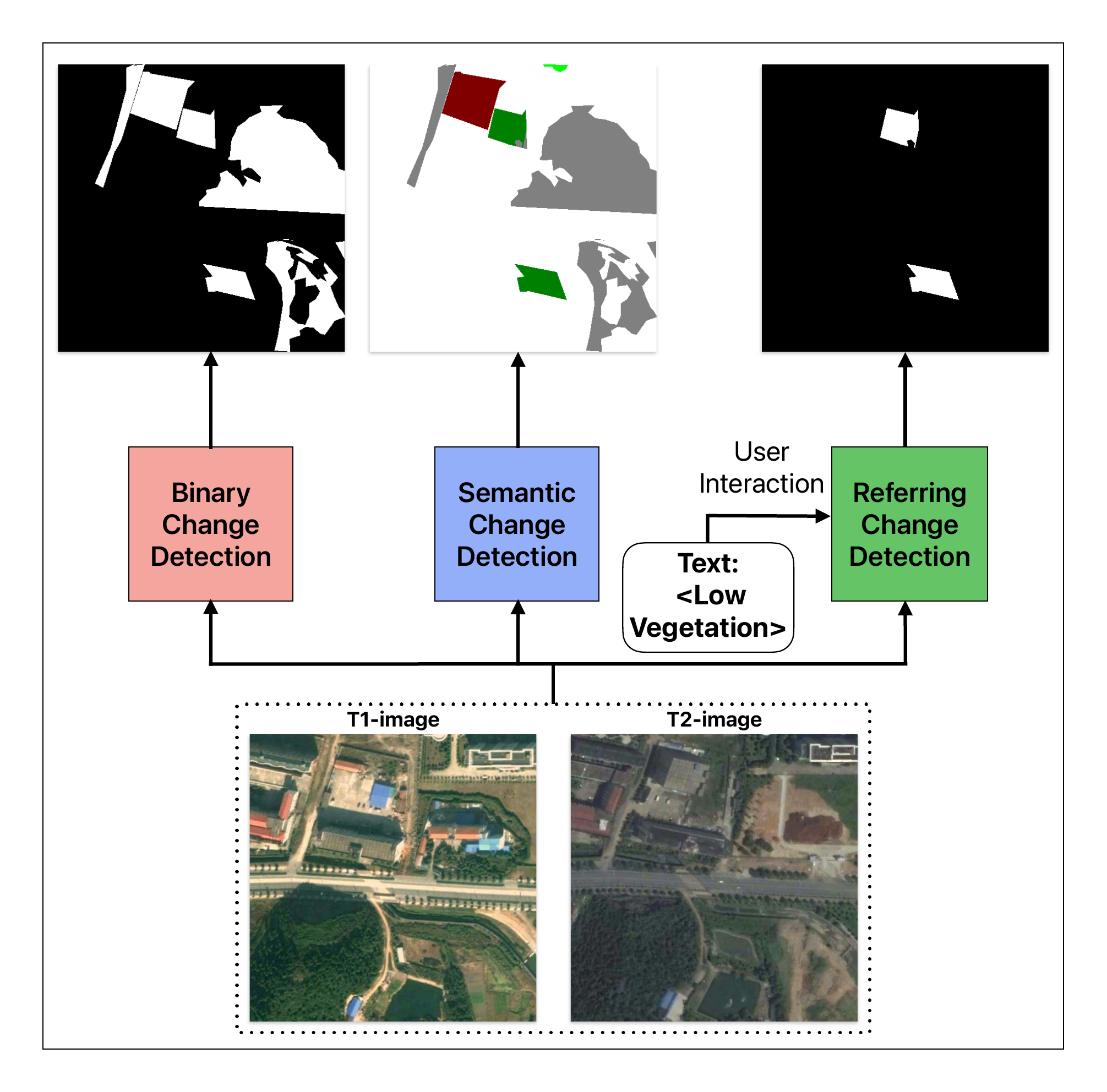}
\vskip-8pt    \caption{Comparison of different approaches in change detection. T1- and T2-images represent distinct time points, corresponding to pre-change and post-change states. Binary change detection methods produce a single map showing all changes that occurred over this time frame, while semantic change detection methods attempt to categorize all changes, both of which limit user interaction. In contrast, referring change detection enables user input to specify particular changes of interest, displaying only those changes that match the user-defined criteria.}
    \label{fig:fig1}
\end{figure}

Change detection (CD) task in remote sensing refers to identifying significant changes to an area between two time intervals. CD is an essential task for urban planning, environmental monitoring and disaster management \cite{hamidi2023fast,he2023cross,haque2023change,coops2023framework}. However, it possesses a significant challenge because remote sensing images taken at different times frequently display data inconsistencies, including illumination changes, noise, and variations in resolution or alignment \cite{liu2014illumination,wan2018illumination,inglada2006analysis}. In recent years deep learning methods have gained significant attraction for CD utilizing convolutional neural networks \cite{fcsiam,snunet,dtscn}, transformers \cite{bit,changeformer}, diffusion \cite{ddpmcd} and more recently state space models \cite{cdmamba}. 

Change detection has traditionally been approached as a binary classification problem on undifferentiated change areas, providing no information about changes across distinct semantic classes. While binary change detection (BCD) offer a general overview of altered areas, many applications require identifying the specific semantic classes associated with these changes. For instance, in urban planning, distinguishing between changes, especially buildings, roads, and vegetation is crucial. Similarly, in environmental monitoring, understanding whether changes involve forest cover, water bodies, or agricultural land provides valuable insights. Several semantic change detection (SCD) methods have been proposed in the literature to provide this detailed information \cite{zhu2024review}. However, SCD methods, which aim to categorize changes into predefined classes, often encounter significant limitations due to inconsistencies in class definitions across different datasets. For instance, in one of the SCD datasets, ``impervious surface" and ``bare ground" are treated as separate classes \cite{zhou2023signet} but they are combined into a single class named "non-vegetated ground surface" in another dataset \cite{yang2020semantic}. These discrepancies make it challenging for SCD models to generalize across datasets, as the same physical features may be classified differently depending on the dataset's labeling schema. The scarcity of SCD datasets intensifies these discrepancies and this lack of standardization leads to problems in accurately detecting and categorizing changes. Because the number of output channels in models corresponds directly to the number of classes, models trained on one dataset struggle to perform effectively on another with differing class definitions. Additionally, training with mixed datasets becomes impractical if class definitions do not align precisely. 

SCD methods face additional challenges due to the limited availability of datasets, which often exhibit significant class imbalances. 
For example, in the SECOND dataset \cite{second}, the non-vegetated ground surface class covers 43\% of the total area of change and appears in 2,689 images, while the ``playground" class represents only 0.38\% and is included in just 129 images. Similarly, in the CNAM-CD dataset \cite{zhou2023signet}, ``water bodies" are present in 394 images, whereas "bare ground" appears in 1,453 images. This class imbalance hinders models from learning and detecting changes in underrepresented classes effectively, limiting the generalization capacity and robustness of SCD methods across diverse real-world applications. 

To address these limitations, we introduce \textbf{Referring Change Detection (RCD)} and propose a language-guided change detection network, \textbf{RCDNet}, that effectively integrates visual and textual inputs to generate accurate targeted change maps. Our RCD task inherently overcomes the constraints of traditional SCD tasks—which rely on predefined classes and a fixed, limited number of output channels—by introducing a flexible, language-guided detection process that allows users to specify semantic changes without predefined classes (see \cref{fig:fig1}). Our approach facilitates the integration of heterogeneous datasets, thereby improving the model’s generalizability and versatility. Furthermore, to support effective training for RCD, we propose a synthetic data generation pipeline, \textbf{RCDGen}, that leverages a latent diffusion model to simultaneously generate realistic post-change images and change maps from a given pre-change image. This allows targeting specific classes in the original image and replacing them with the desired post-change category, thus allowing us to produce a much wider spectrum of post-change images than what is available in existing datasets. Our contributions can be summarized as:
\begin{itemize}
    \item We define Referring Change Detection (RCD) task, enabling user interaction in change detection through natural language inputs. 
    \item We propose a synthetic data generation pipeline that simultaneously outputs post-change image and change map given a pre-change image and desired change category.
    \item We introduce a lightweight and efficient RCD model designed to capture only user-intended changes that can be also be adapted to SCD and BCD tasks.

\end{itemize}

\section{Related Work}

\textbf{Traditional change detection methods: } Change Detection has largely been studied as a binary segmentation problem, where the objective is to compute one change map for a given image pair. Usually, the change map of interest depends on the object of interest defined by the dataset. For instance, LEVIR-CD \cite{levir} only provides change maps for buildings. Classical CD methods involve using algebraic algorithms to compute change maps \cite{alge1,alge2,alge3}, transformative approaches which generate transforms on the pre- and post-images followed by a difference \cite{pca1,pca2,mad,irmad}. With the advent of Deep Learning (DL), change detection has also moved from classical methods to DL-based methods. Convolutional Neural Networks (CNNs) have been widely used in CD, including FC-EF \cite{fcsiam}, FC-Siam \cite{fcsiam}, SNUNet \cite{snunet}, IFNet \cite{ifnet}, and DT-SCN \cite{dtscn}. 

An alternative to CNN-based methods is the transformer-based methods, which include BiT \cite{bit}, STANet \cite{stanet}, and ChangeFormer \cite{changeformer}. These utilize the Vision Transformer (ViT) \cite{dosovitskiy2020image}, due to its ability to capture long-range dependencies better than CNNs. On the other hand, DDPM-CD \cite{ddpmcd} uses generative modeling to learn powerful encoders. Very recently, selective state space models (S4), i.e., Mamba, have been used instead of CNNs or transformers for CD following its computational efficiency \cite{gu2023mamba}. These methods include RSMamba \cite{rsmamba}, ChangeMamba \cite{changemamba}, CDMamba \cite{cdmamba} and M-CD \cite{m-cd}. Our approach leverages the Mamba architecture for its computational efficiency, as in prior work, and improves it with cross-attention transformers to enable multimodality via language guidance, resulting in a Mamba–Transformer hybrid architecture.

\textbf{Semantic change detection methods: } The changes related to different objects of interest can vary significantly depending on the application. Therefore, generating semantic change maps with multiple classes, rather than a simple binary map, is more valuable. As a result, several studies have aimed to address the task of Semantic Change Detection (SCD). Gousseau et al. \cite{daudt2019multitask} are among the first to publish a dataset for semantic change detection (SCD) and propose a fully convolutional network for the task. Zhang et al. \cite{second} also release a dataset, named SECOND, and introduce asymmetrical siamese networks as a solution. In \cite{ding2022bi}, the authors present an innovative CNN architecture for SCD called SSCD-1, where semantic temporal features are integrated within a deep change detection unit; they later propose a more efficient model named Bi-SRNet, which incorporates a novel semantic consistency loss. In \cite{ding2024joint}, the authors introduce a CNN-based Triple Encoder-Decoder network (TED) for extracting temporal features and change representations, followed by a Semantic Change Transformer (SCanFormer) designed to explicitly capture semantic transitions in token space. They then present SCanNet, a hybrid ‘CNN-Transformer’ method that uses TED for feature extraction and SCanFormer to capture correlations within the embedded semantic space.

\textbf{LLM-based Methods in Remote Sensing:} Various concurrent works have utilized foundation models for change detection and other geo-sensing-based tasks. One of these works called SeFi-CD \cite{sefi} introduces a framework where text prompts are used to generate visual prompts for frozen foundation models which further generate the change map. This method comes closest to our approach. However, a major difference in our method is that we utilize the foundation model CLIP \cite{clip} for generating the text embedding, but train a Mamba-based encoder and decoder instead of using a foundation model for encoding the image. The reason behind that is the observation that the available foundation models are not trained heavily with remote-sensing images and might hallucinate. Since implementation is not public at the time of submission we are not able to include this method in comparison. Another concurrent work called CDChat \cite{cdchat} tries to generate a text description of the change between two temporal images using Vicuna \cite{vicuna}. In addition, it can also perform visual question answering based on the input pre- and post-change images. However, this method does not generate visual change maps. A recent work, BAN \cite{li2024new}, repurposes CLIP's ViT for bi-temporal change detection through a Bi-Temporal Adapter Network (BAN) with bridging modules that inject general features into a task-specific CD branch; however, it is trained in either the BCD or SCD setting, which keeps our approach distinct.

\textbf{Synthetic Datasets in Remote Sensing: }
Song et al. \cite{song2024syntheworld} present a Stable Diffusion and 3D modeling based synthetic image generation pipeline that uses GPT-4 for prompt creation, allowing images to be generated entirely from scratch. However, their change detection dataset exclusively captures building changes, resulting in the absence of other semantic classes. Meanwhile, Tang et al. \cite{tang2024changeanywhere} propose a Stable Diffusion based generation pipeline that depends on semantic segmentation masks during inference, thus limited with existing segmentation datasets. In a similar vein, Zheng et al. \cite{zheng2023scalable} introduce a GAN based synthetic image generator that incrementally adds or removes objects from pre-event images, also relying on annotated segmentation data. While these methods can yield highly realistic simulations, they differ from our approach in that we do not require semantic segmentation masks during inference. Building on \cite{zheng2023scalable}, Zheng et al. \cite{zheng2024changen2} propose a self-supervised technique and latent diffusion based model, removing the need for pre-event annotations by using object contours as diffusion conditions. While this innovation reduces annotation requirements, it relies on off-the-shelf foundation models (e.g., SAM \cite{kirillov2023segment}) to generate the contours at inference time, potentially limiting performance based on these models’ capabilities.

\section{Methodology}
The RCD task can be defined similar to the BCD task, albeit with an additional user-input selecting the change of interest. Thus, the goal of RCD is to learn a function \(f\) that satisfies \(M = f(I_{pre}, I_{post}, C)\) where \(I_{pre}\) and \(I_{post}\) denote the pre- and post-change images respectively, and \(M\) denotes the change map. The user intent is specified by \(C\). In our method, we use CLIP-based text embeddings of the change category to model the user intent. Next, we describe our RCDGen pipeline and RCDNet architecture.

\subsection{RCDGen}
\begin{figure*}
    \centering
    \includegraphics[width=.9\textwidth]{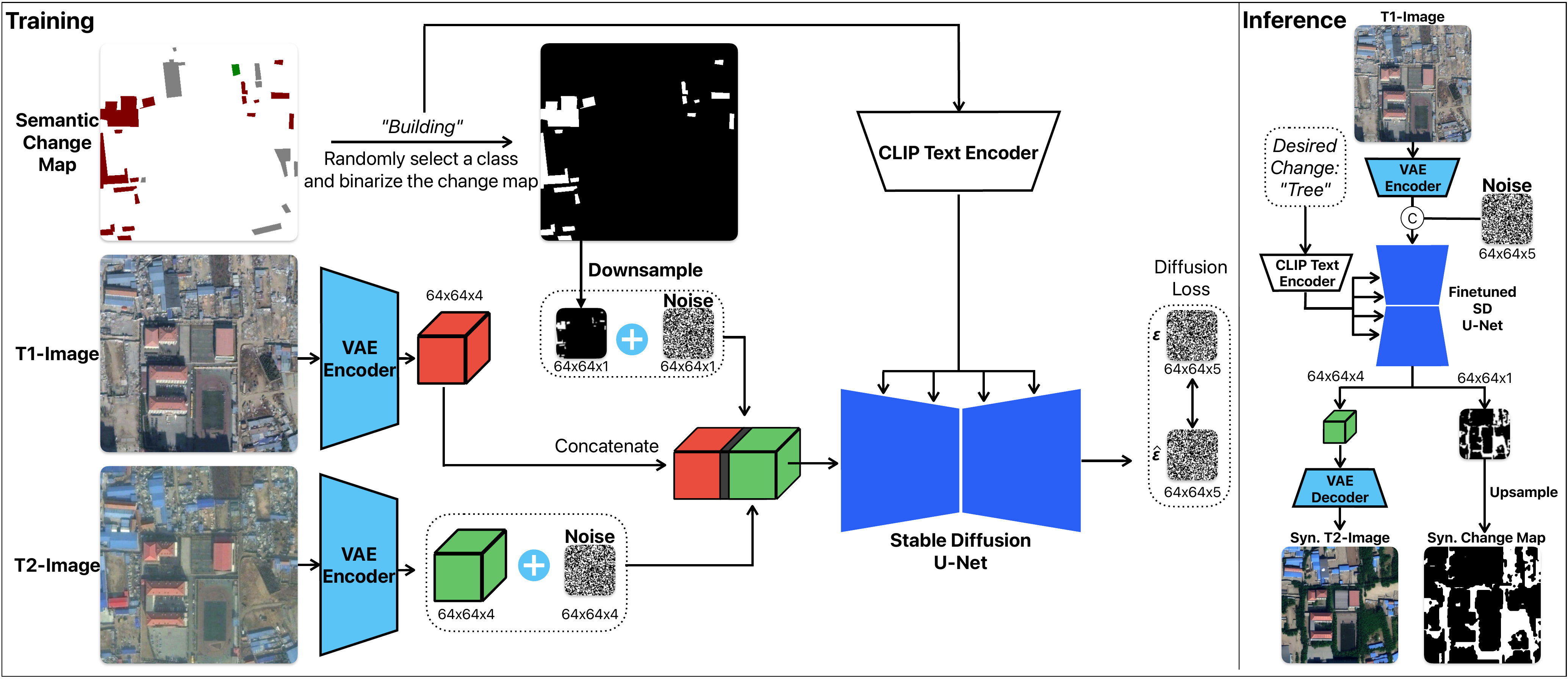}
\vskip-9pt    \caption{The training and inference scheme of RCDGen is illustrated. In the training phase, the pre-change latent is shown in red, and the post-change latent in green. The change maps are downsampled to match the resolution of the latents ($64 \times 64$) to enable concatenation. During inference, pre-change latent is concatenated with noise of shape $64 \times 64 \times 5$. After denoising, the generated latent is split back to match the channels used during training. The generated downsampled mask is upsampled via interpolation, and the generated latent is decoded via the VAE Decoder.}
    \label{fig:syn_train}
\end{figure*}

\begin{figure*}
    \centering
    \includegraphics[width=0.88\linewidth]{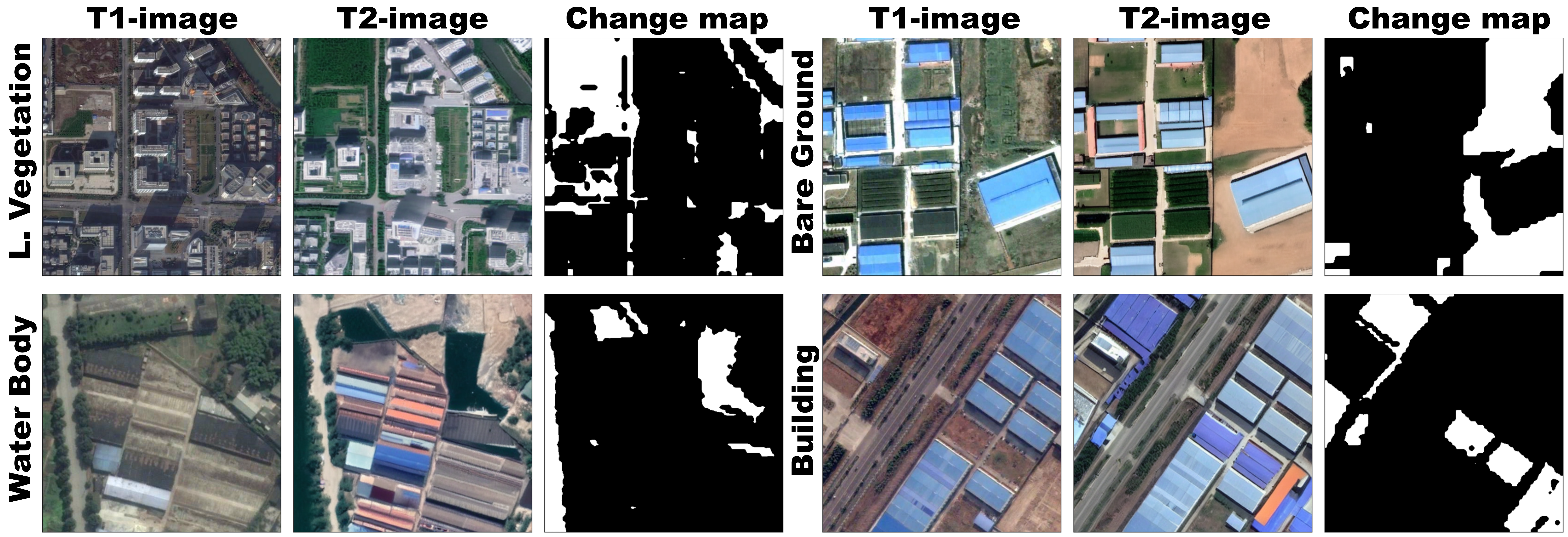}
   \vskip-10pt \caption{Synthetic samples generated by our pipeline: T1-images (pre-change) are sourced from real datasets, while T2-images (post-change) and corresponding change maps are generated. Labels on the left side of each image pair indicate the specific change category.} 
    \label{fig:synthetic_samples}
\end{figure*}

Semantic change detection datasets are often small in scale and contain limited semantically annotated examples, often accompanied by severe class imbalance, making it challenging to train robust models and increasing the risk of overfitting. To address these limitations for RCD, we propose RCDGen, a generative pipeline that synthesizes post-change images conditioned on existing pre-change images and target classes, allowing us to produce any number of post-change images with the desired class, ensuring a well-balanced and diverse dataset. We illustrate the RCDGen's training and inference procedures in \cref{fig:syn_train}. We modify the InstructPix2Pix pipeline \cite{brooks2023instructpix2pix} and utilize existing SCD datasets to learn to model the joint distribution of post-change images and change maps conditioned on pre-change images. More specifically, we first use a Variational Auto Encoder (VAE) to generate latents for the pre- and post-change images. Then, we randomly select a change class and generate a change map for this class by masking other pixels in the annotated change map, followed by downsampling to match the shape of the latents. Both the latents and the change map are concatenated and passed through the forward diffusion process, where random noise is added to only the post-image latent and the change map. During the denoising process, we utilize the CLIP text embeddings of the class name to guide the UNet to predict the noise added during the forward process. This setup is trained using the canonical diffusion loss \cite{ho2020denoising} denoted as follows:

\begin{equation}
\begin{aligned}
L = \mathbb{E}_{\epsilon \sim \mathcal{N}(0,1)} \Big[ \Big\| \epsilon - \epsilon_\theta(z_t, t, \mathcal{E}(T1), c_T) \Big\|_2^2 \Big],
\end{aligned}
\end{equation}
where $\mathcal{E}$ denotes VAE encoder, $\epsilon_\theta$ denotes UNet, $c_T$ represents the text embeddings, $t$ is a random timestep and $z_t$ is defined as:
\begin{equation}
    z_t = \sqrt{\alpha_t}\times[\mathcal{E}(T2) \oplus \tilde{m}] + \sqrt{1-\alpha_t}\times\epsilon,
\end{equation}
where $\oplus$ denotes concatenation in channel dimension, $\alpha_t$ corresponds to the noise schedule term, $\epsilon$ denotes the added noise, and $\tilde{m}$ is the downsampled change map.

For training, We utilize images from the training sets of the SECOND \cite{second}, CNAM-CD \cite{zhou2023signet}, LEVIR-CD \cite{levir} and LEVIR-CD+ \cite{levir} datasets. For inference, InstructPix2Pix \cite{brooks2023instructpix2pix} extends the classifier-free guidance \cite{ho2022classifier} sampling strategy by incorporating dual conditioning inputs—an image and an instruction—each applied with separate coefficients to control their respective strengths. We adapt this strategy to balance the desired change with consistency to the pre-image. The modified noise estimation is given below:
\begin{equation}
\begin{aligned}
\tilde{e_\theta}&\left(z_t, \mathcal{E}(T1), c_T\right)=  e_\theta\left(z_t, \varnothing, \varnothing\right) \\
&+s_I \cdot\left(e_\theta\left(z_t, \mathcal{E}(T1), \varnothing\right)-e_\theta\left(z_t, \varnothing, \varnothing\right)\right) \\
& +s_T \cdot\left(e_\theta\left(z_t, \mathcal{E}(T1), c_T\right)-e_\theta\left(z_t, \mathcal{E}(T1), \varnothing\right)\right),
\end{aligned}
\label{eq:inference_syn}
\end{equation}
where $\varnothing$ represents the null condition, $\tilde{e_\theta}$ denotes the modified estimate, \( s_I \) controls the strength of the image condition, and \( s_T \) controls the strength of the text condition. 

To create a more comprehensive dataset, we additionally incorporate a relatively small number of inpainted samples within our synthetic dataset, generated by adapting stable diffusion for inpainting. These inpainted samples introduce subtle, realistic changes that complement the larger transformations in the primary synthetic data, achieving a more balanced representation of change types. For this inpainting process, we leverage an existing remote sensing semantic segmentation dataset \cite{wang2021loveda}, where specific classes are randomly masked using semantic segmentation masks and modified based on prompts, following a similar pipeline but without the InstructPix2Pix sampling scheme. This combination of subtle and significant changes allows our synthetic dataset to better capture a diverse range of transformations. Our final synthetic dataset includes 29,500 images synthesized from SECOND, 17,500 from CNAM-CD, and 8,300 inpainted samples, creating a robust and varied resource for referring change detection. We utilize this synthetic dataset for pretraining of the RCDNet. We show sample synthetic images generated by our pipeline in \cref{fig:synthetic_samples}. A more comprehensive set of samples are provided in the supplementary document. We believe this dataset and generation pipeline would be highly beneficial for the community.

\begin{figure*}[htp]
    \centering
    \includegraphics[width=.95\linewidth]{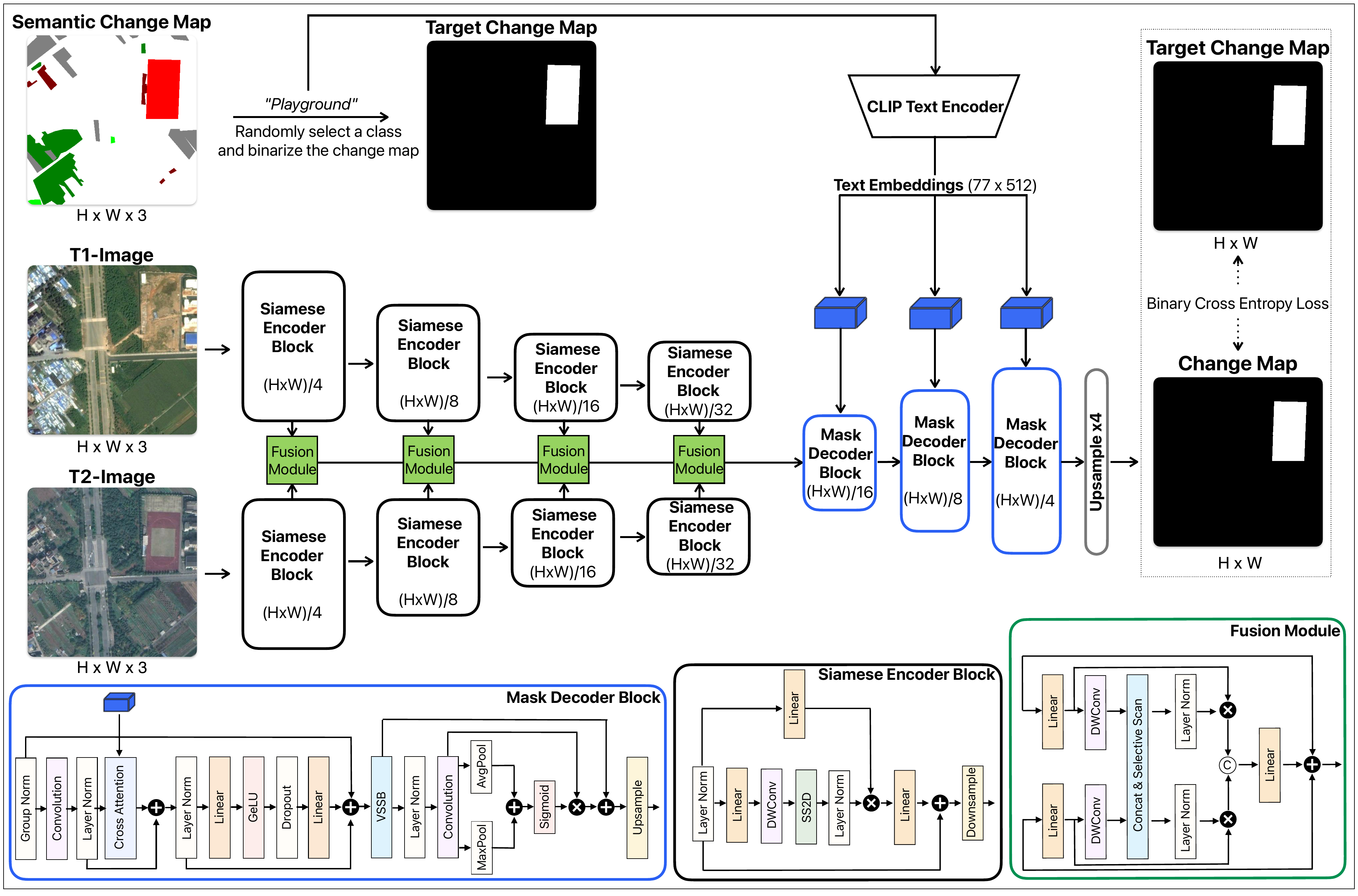}
\vskip-8pt    \caption{The RCDNet and its training scheme are illustrated. T1- and T2-images represent distinct time points, corresponding to pre-change and post-change states. The change category (class) is randomly selected from unique values in the semantic change map and guides the network through text embeddings. The change map is binarized for that specified category, enabling the use of canonical binary cross-entropy loss as training objective.}
    \label{fig:architecture}
\end{figure*}

\subsection{RCDNet}
We illustrate the architecture of our proposed model in \cref{fig:architecture}. Our design is primarily inspired by the state-of-the-art Mamba-based semantic segmentation model \cite{wan2024sigma}, which has demonstrated strong performance in binary change detection (BCD) tasks \cite{m-cd}. Specifically, we use a VMamba-based \cite{liu2024vmamba} Siamese Encoder Block (SEB) to process both the pre-change (T1-image) and post-change (T2-image) images in parallel, extracting meaningful features from each. These features are then passed to a Fusion Module (FM), which performs a comparative analysis and produces fused latent representations. Finally, these fused latents are input to a series of lightweight Mask Decoder Blocks (MDB), which generate the change map conditioned on user intent, provided as text embeddings corresponding to the desired change class. We detail the key components of RCDNet in the following sections.

\noindent\textbf{Siamaese Encoder Block (SEB): }
Siamese Encoder Block is built with Visual State Space Blocks (VSSB) \cite{liu2024vmamba}, as shown with black frame in \cref{fig:architecture}. A series of consecutive SEBs forms an encoder network featuring two identical weighted branches (Siamese architecture) that process both T1- and T2-images, representing distinct time points. In each SEB, the resolution of feature maps is halved while the number of channels is doubled, enhancing feature representation. Each encoded feature is directed to separate fusion modules, enabling the selection of differential features to be shared with the Mask Decoder Blocks.


\noindent\textbf{Fusion Module (FM): }
The fusion module receives features from the two branches and produces a single feature vector for the mask decoder for each resolution, shown with green frame in \cref{fig:architecture}. FM blocks consists of linear, depth-wise convolution and Concat \& Selective Scan (CSS) modules, following \cite{wan2024sigma} and \cite{m-cd}. In the CSS module, the two features are initially flattened and then concatenated along the sequence length dimension. Here, the concatenation of feature vectors from both branches allows selective scan module to process both feature vectors together, enhancing the cross-branch long range contextual sensitivity unlike relying solely on convolution layers for feature fusion.


\noindent\textbf{Mask Decoder Block (MDB): }
The Mask Decoder Blocks, highlighted with blue in \cref{fig:architecture}, are responsible for generating binary change maps based on text embeddings using cross-attention transformer layers. To enhance the learning of inter-channel dependencies, we incorporate both average-pooling and max-pooling operations on feature channels, following \cite{wan2024sigma}. The CLIP Text Encoder generates text embeddings using the target class names as inputs. These text embeddings serve as keys and values in the cross-attention transformer, guiding the selection of visual queries.



\section{Experiments and Results}
To enable fair comparison against existing SCD and BCD methods, we use only the corresponding change category (class name) as the text prompt and conduct extensive experiments, spanning from classical SCD to cross-domain tests. In all tables, cases where the synthetic data is used for pretraining are indicated in parentheses. For within-domain scenarios, all models are trained zsolely on the target dataset. For the cross-domain cases, all SCD methods are trained exclusively on the SECOND dataset, as it includes buildings as a sub-class.

\begin{table*}
\centering
\caption{Within-domain and cross-domain performance of SCD methods.}
\resizebox{\textwidth}{!}{
\begin{tabular}{@{}lcccccccc||cccc@{}}
\toprule
\textbf{Method} & \multicolumn{4}{c}{\textbf{SECOND} \cite{second}} & \multicolumn{4}{c}{\textbf{CNAM-CD} \cite{zhou2023signet}} & \multicolumn{2}{c}{\textbf{WHU-CD} \cite{ji2018fully}} & \multicolumn{2}{c}{\textbf{LEVIR-CD} \cite{levir}} \\ 
\cmidrule(lr){2-5} \cmidrule(lr){6-9} \cmidrule(lr){10-11} \cmidrule(lr){12-13}
 & \textbf{mIoU} (\(\uparrow\)) & \textbf{SeK} (\(\uparrow\)) & \textbf{OA} (\(\uparrow\)) & {\boldmath$\mathrm{F_{scd}}$} (\(\uparrow\)) & \textbf{mIoU} (\(\uparrow\)) & \textbf{SeK} (\(\uparrow\)) & \textbf{OA} (\(\uparrow\)) & {\boldmath$\mathrm{F_{scd}}$} (\(\uparrow\)) & \textbf{IoU} (\(\uparrow\)) & \textbf{OA} (\(\uparrow\)) & \textbf{IoU} (\(\uparrow\)) & \textbf{OA} (\(\uparrow\)) \\ \midrule
\textbf{Experiment} & \multicolumn{4}{c}{within-domain} & \multicolumn{4}{c}{within-domain} & \multicolumn{2}{c}{cross-domain} & \multicolumn{2}{c}{cross-domain} \\ \midrule
HRSCD-str.3 \cite{daudt2019multitask} & 66.74 & 13.21 & 84.63 & 56.51 & 57.36 & 0.62 & 77.24 & 28.38 & 43.75 & 96.04 & 42.74 & 95.74 \\ \midrule
BAN \cite{li2024new} & 66.55 & 10.87 & 84.59 & 53.15 & 62.51 & 2.63 & 77.25 & 35.76 & 29.78 & 93.23 & 18.71 &91.30  \\ \midrule
HRSCD-str.4 \cite{daudt2019multitask} & 72.24 & 20.56 & 87.60 & 63.21 & 67.94 & 3.31 & 78.94 & 34.14 & 43.97 & 95.90 & 48.29 & 96.40 \\ \midrule
SSCD-1 \cite{ding2022bi} & 73.03 & 22.50 & 88.07 & 65.36 & 71.17 & 13.47 & 82.08 & 46.80 & 46.14 & 96.05 & 49.64 & 96.01 \\ \midrule
Bi-SRNet \cite{ding2022bi} & 73.04 & 22.26 & 88.19 & 65.21 & 69.11 & 19.22 & 82.25 & 58.55 & 46.57 & 96.27 & 44.74 & 95.28 \\ \midrule
TED \cite{ding2024joint} & \textbf{73.53} & 22.90 & 88.14 & 65.28 & 67.00 & 9.65 & 78.98 & 44.78 & 44.46 & 95.69 & 51.43 & 96.11 \\ \midrule
SCanNet \cite{ding2024joint} & 73.42 & 23.90 & 88.20 & 66.66 & 71.38 & 22.68 & 84.69 & 61.21 & 38.32 & 94.58 & 47.29 & 95.29 \\ \midrule
RCDNet & 73.04 & 24.57 & 89.00 & 68.29 & 72.83 & 29.49 & 87.35 & 70.41 & 59.05 & 97.73 & 53.98 & 96.78 \\ \midrule
RCDNet (synthetic) & 73.47 & \textbf{25.25} & \textbf{89.24} & \textbf{68.90} & \textbf{75.32} & \textbf{33.81} & \textbf{88.95} & \textbf{73.50} & \textbf{63.14} & \textbf{97.99} & \textbf{60.21} & \textbf{97.50} \\ \bottomrule
\end{tabular}}
\label{tab:combined_performance}
\end{table*}

\subsection{Implementation Details}
We train our model using the AdamW optimizer \cite{loshchilov2017decoupled} with an initial learning rate of \(6 \times 10^{-5}\) and a weight decay of 0.01. The learning rate follows a polynomial decay schedule with an exponent of 0.9. Training is conducted over 200 epochs with a batch size of 4 with an NVIDIA RTX A6000 GPU with 48GB of memory. For the synthetic data pipeline, we fine-tune stable diffusion for 15,000 iterations using a batch size of 4. Fine-tuning is carried out with the Hugging Face Diffusers library \cite{von-platen-etal-2022-diffusers} and leverages the pretrained InstructPix2Pix checkpoint \cite{brooks2023instructpix2pix}. During synthetic data generation, we set \( s_I \) to 1.5 and \( s_T \) to 7.0.

\subsection{Datasets}
We use the benchmark SCD datasets SECOND \cite{second} and CNAM-CD \cite{zhou2023signet} for within-domain comparisons for the SCD tasks. For cross-domain experiments and comparison with BCD methods, we employ the BCD datasets LEVIR-CD \cite{levir} and WHU-CD \cite{ji2018fully}. Detailed description of each dataset is provided in the supplementary document.

\subsection{Evaluation Metrics}
We evaluate the performance using four standard metrics for semantic change detection: overall pixel accuracy (OA), semantic mean intersection over union (mIoU) \cite{second}, separated kappa coefficient (SeK) \cite{second}, and semantic change detection F1 score ($\mathrm{F_{scd}}$) \cite{ding2022bi}. Detailed definitions are provided in the supplementary document.

\subsection{Within-domain in SCD}
\begin{figure}
    \centering
    \includegraphics[width=1\linewidth]{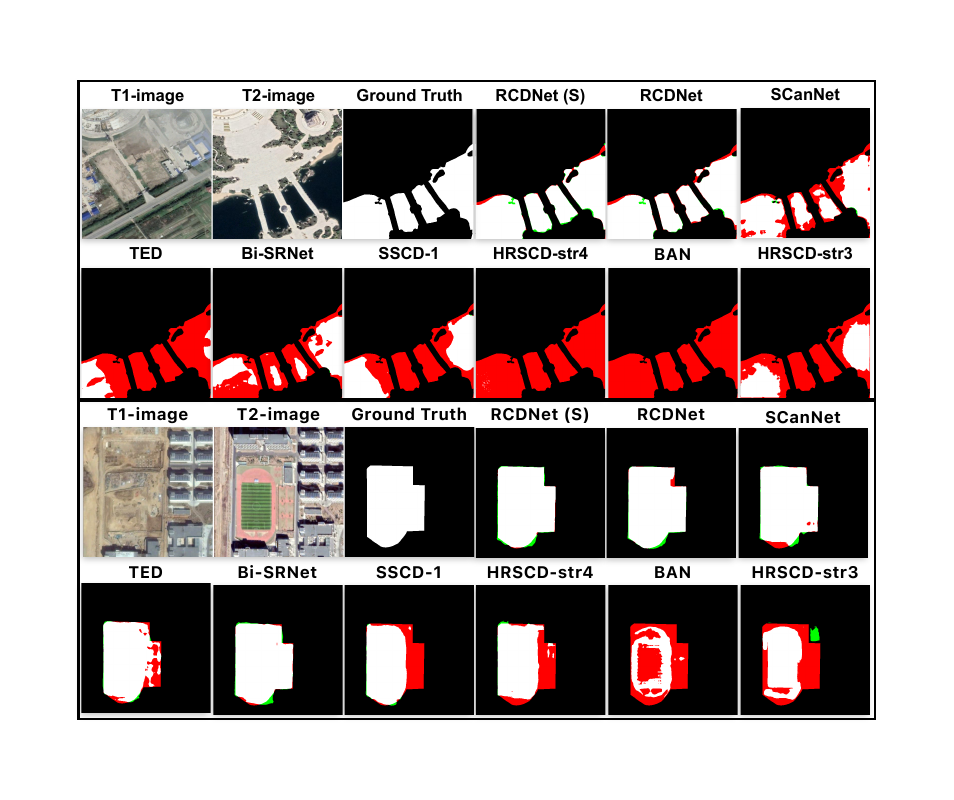}
\vskip-8pt    \caption{Qualitative results corresponding to the "Water Bodies" category in CNAM-CD (upper row) and "Playground" category in SECOND (lower row) datasets. T1-images represent pre-change (earlier time point) and T2-images represent post-change (later time point). In the results, white pixels indicate true positives, black pixels indicate true negatives, green pixels denote false positives, and red pixels denote false negatives. The notation "(S)" refers to synthetic pretraining applied to RCDNet.}
    \label{fig:combined_rep}
\end{figure}

We evaluate our proposed model against state-of-the-art SCD methods, conducting within-domain tests where the training and test sets originate from the same SCD dataset. Although such aggregation is unnecessary in real-world settings, to align RCD results with SCD we combine the class-specific binary change maps, obtained from multiple forward passes, into a unified semantic change map for each image. We focus on post-change maps (i.e., growth changes) for both datasets. Qualitative results, as shown in \cref{fig:combined_rep}, illustrate the model’s ability to capture precise, user-specified changes, highlighting the added detail and accuracy enabled by our approach. The left part of \cref{tab:combined_performance} presents within-domain performance for SECOND and CNAM-CD, where RCDNet demonstrates superior performance with or without synthetic data, outperforming the second-best methods by notable margins: on SECOND, it surpasses in SeK by 1.35, OA by 1.04, and $\mathrm{F_{scd}}$ by 2.24; on CNAM-CD, it achieves an additional 3.94 in mIoU, 11.13 in SeK, 4.26 in OA, and 12.29 in $\mathrm{F_{scd}}$. These quantitative and qualitative improvements underscore the advantages of RCD’s targeted approach, where user-defined change categories enable RCDNet to focus on specific, contextually relevant transformations. This targeted approach, combined with synthetic data pretraining, delivers more nuanced and accurate results than standard SCD methods, which broadly classify all changes without customization for user needs.

\subsection{Within-domain in BCD}
We evaluate performance on building change detection using the single-class BCD datasets LEVIR-CD and WHU-CD. Given the single-class focus, SCD metrics are not applicable; thus, we rely on the original BCD metrics: IoU and overall pixel accuracy (OA). \cref{tab:comparison_whu_levir} show within-domain performance comparison for LEVIR-CD and WHU-CD datasets. On WHU-CD, RCDNet (synthetic) outperforms the state-of-the-art best IoU (M-CD \cite{m-cd}) by 0.8 and matches the top OA at 99.7. On LEVIR-CD, it achieves a 0.8 improvement in IoU over M-CD \cite{m-cd} and shares the highest OA of 99.2. These results demonstrate RCDNet’s state-of-the-art performance in the BCD task, with no performance drop despite the focus on user-specified categories, underscoring the robustness of RCDNet’s targeted approach and its adaptability to BCD along with the SCD tasks. 

\begin{table}[h]
\centering
\caption{Mixed dataset performance in SCD datasets.}
\small
\resizebox{.75\columnwidth}{!}{
\begin{tabular}{@{}lcc@{}}
\toprule
\textbf{Metric} & \textbf{RCDNet (mixed)} & \textbf{RCDNet (distinct)} \\ \midrule
\multicolumn{3}{c}{\textbf{SECOND} \cite{second}} \\ \midrule
mIoU (\(\uparrow\)) & 72.98 & \textbf{73.04} \\
SeK (\(\uparrow\)) & 24.48 & \textbf{24.57} \\
OA (\(\uparrow\)) & \textbf{89.27} & 89.00 \\
$\mathrm{F_{scd}}$ (\(\uparrow\)) & 68.26 & \textbf{68.29} \\ \midrule
\multicolumn{3}{c}{\textbf{CNAM-CD} \cite{zhou2023signet}} \\ \midrule
mIoU (\(\uparrow\)) & \textbf{73.39} & 72.83 \\
SeK (\(\uparrow\)) & \textbf{30.00} & 29.49 \\
OA (\(\uparrow\)) & \textbf{87.51} & 87.35 \\
$\mathrm{F_{scd}}$ (\(\uparrow\)) & \textbf{70.61} & 70.41 \\ \bottomrule
\end{tabular}}
\vskip-6pt
\label{tab:mixed_performance}
\end{table}

\subsection{Cross-domain Capabilities}
We evaluate the performance in a cross-domain scenario, where the target class (building) is familiar to the model through training on the SECOND dataset, but the test images are drawn from the WHU-CD and LEVIR-CD datasets. The right part of \cref{tab:combined_performance} shows the cross-domain performance for the LEVIR-CD and WHU-CD datasets. Results shows that RCDNet with synthetic data significantly outperforms the other methods in cross-domain performance on the WHU-CD and LEVIR-CD datasets. Compared to the next best methods, it achieves a notable IoU improvement of 16.57 on WHU-CD and 8.78 on LEVIR-CD, along with higher OA scores. Additionally, synthetic data boosts RCDNet’s performance over its non-synthetic version by 4.09 IoU on WHU-CD and 6.23 IoU on LEVIR-CD, highlighting the effectiveness of synthetic data in enhancing RCDNet’s adaptability across datasets. The diversity introduced through our synthetic data generation pipeline plays a key role in enabling this benefit, as it exposes RCDNet to a wider range of visual variations and change scenarios during pretraining. By simulating diverse instances of the target class across various contexts and conditions, the synthetic data allows RCDNet to learn more generalized feature representations. This improved representation enables the model to handle previously unseen imagery more effectively, capturing complex patterns and subtle changes that might otherwise be missed. We also experiment in a completely zero-shot setting where the target change category is unknown to the model. While some results are promising, performance is not consistently reliable across different images and categories. Nevertheless, if the text embedding of an unseen category is sufficiently similar to those of seen categories, zero-shot transfer should in principle be feasible. Due to these inconsistent results, we refrain from making strong claims and leave this as a promising direction for future research.

\begin{table}
\centering
\caption{Within-domain performance in BCD datasets.}
\resizebox{1\columnwidth}{!}{
\begin{tabular}{@{\extracolsep{4pt}}c c c c c c@{}}
\toprule
 & \multicolumn{2}{c}{WHU-CD \cite{ji2018fully}} & \multicolumn{2}{c}{LEVIR-CD \cite{levir}}\\
\cline{2-3} \cline{4-5} \\
Method & \textbf{IoU} (\(\uparrow\)) & \textbf{OA} (\(\uparrow\)) & \textbf{IoU} (\(\uparrow\)) & \textbf{OA} (\(\uparrow\)) \\
\midrule
FC-Siam-conc \cite{fcsiam} & 66.5 & 98.5 & 72.0 & 98.5 \\
SNUNet \cite{snunet} & 71.7 & 98.7 & 78.8 & 98.8 \\
IFNet \cite{ifnet} & 71.5 & 98.8 & 78.8 & 98.9 \\
DT-SCN \cite{dtscn} & 84.2 & 99.3 & 78.1 & 98.8 \\
STANet \cite{stanet} & 70.0 & 98.5 & 77.4 & 98.7 \\
BIT \cite{bit} & 83.4 & 99.3 & 80.7 & 98.9 \\
ChangeFormer \cite{changeformer} & 79.5 & 99.1 & 82.5 & 99.0 \\
ChangeMamba \cite{changemamba} & 86.1 & 99.4 & 82.1 & 99.0 \\
CDMamba \cite{cdmamba} & 88.2 & 99.5 & 83.1 & 99.0 \\
DDPM-CD \cite{ddpmcd} & 86.3 & 99.4 & 83.3 & 99.1 \\
M-CD \cite{m-cd} & 91.1 & 99.6 & 85.0 & \textbf{99.2} \\
RCDNet  & 91.6& \textbf{99.7} & \textbf{85.8} & \textbf{99.2}\\
RCDNet (synthetic) &\textbf{91.9} & \textbf{99.7}& \textbf{85.8} & \textbf{99.2}
\\ 
\bottomrule
\end{tabular}}
\vskip-10pt

\label{tab:comparison_whu_levir}
\end{table}

\subsection{Mixed Dataset Performance}
To evaluate the impact of combining datasets on the performance of our RCDNet model, we conduct experiments using a merged dataset. This approach consolidates multiple datasets into a single training phase, eliminating the need for separate training sessions for each dataset and thus streamlining the overall process. For comparison, we combine the SECOND and CNAM-CD datasets and train our model on this mixed dataset. The performance metrics are shown in \cref{tab:mixed_performance}, indicating that any performance degradation is negligible compared to training on individual datasets and even surpasses performance on CNAM-CD. These results confirm that our approach effectively preserves model accuracy and robustness, even when trained on a diverse mix of datasets, overcoming a key limitation of traditional SCD methods, which cannot be effectively trained with combined datasets due to their strict requirement for matching class numbers and names.

\subsection{Ablation Studies}  
We perform extensive ablation studies to refine our architecture, using a composite evaluation metric of $0.3 \times \text{mIoU} + 0.7 \times \text{SeK}$ following \cite{zhou2023signet} and report this score in SECOND validation set. Key findings are summarized below:
\begin{enumerate}
    \item \textbf{Encoder Backbone:} We evaluate several encoder backbones of different sizes, including Swin Transformer-B (medium), Swin Transformer-S (small) \cite{liu2021swin}, Segformer-b2 (small), Segformer-b5 (large) \cite{xie2021segformer}, and the VMamba series in tiny, small, and base configurations \cite{liu2024vmamba} (see supplementary document for comparison graph). Based on these tests, we select VMamba-small as our encoder, which yielded the best results, confirming the observations of \cite{m-cd}.
    
    \item \textbf{CLIP Variants:} To assess the value of domain-specific knowledge, we compare the remote sensing-specific RemoteCLIP model \cite{liu2024remoteclip} with OpenAI’s general CLIP models \cite{clip}. Results show no advantage of RemoteCLIP over the original CLIP weights, highlighting the strong generalization capabilities of OpenAI’s CLIP, which has been pretrained on a very extensive and diverse dataset. (see supplementary document for comparison). 
    
    \item \textbf{Pretraining Dataset:} Models are pretrained on synthetic data, ImageNet, or initialized from scratch. Pretraining on synthetic data provides a substantially stronger initial performance and better overall performance. See the supplementary document for a comparison over training epochs and performance advantage in SCD and BCD tasks provided by synthetic dataset pretraining.
    
    \item \textbf{Decoder Backbone:} Both MLP-based and Mamba-based decoders are evaluated. The Mamba-based decoder outperforms the MLP-based option with comparable computational expense. More details are available in the supplementary document.
    
    \item \textbf{CLIP Freezing:} We assess the impact of freezing versus fine-tuning the CLIP weights using LORA \cite{hu2021lora}.  Supplementary document includes performance comparisons for both cases.
\end{enumerate}

\section{Conclusion}
We introduce RCD, a flexible approach for user-directed change detection in remote sensing imagery. RCD addresses the limitations of traditional binary and semantic change detection methods by integrating pre-trained language models with a change detection network, allowing users to specify desired detection categories via natural language prompts. Additionally, our diffusion-based synthetic data generation pipeline produces realistic post-change images and change maps, mitigating data scarcity and class imbalance to improve overall and cross-domain performance of RCDNet, our proposed RCD model.
\section{Acknowledgement}
This work was supported by the NSF CAREER award under Grant 2045489 and by the Army Research Laboratory under Cooperative Agreement Number W911NF-23-2-0008.  The views and conclusions contained
in this document are those of the authors and should not be interpreted as representing the official policies, either expressed or implied, of the Army Research Laboratory or the U.S. Government. The U.S. Government is authorized to reproduce and distribute
reprints for Government purposes notwithstanding any copyright notation herein.
{
    \small
\bibliographystyle{ieeenat_fullname}
    \bibliography{main}
}

\end{document}